\documentclass{article}

\usepackage{arxiv}

\usepackage{times}
\usepackage{soul}
\usepackage{url}
\usepackage{hyperref}
\usepackage[utf8]{inputenc}
\usepackage[small]{caption}
\usepackage{graphicx}
\usepackage{amsmath}
\usepackage{amsthm}
\usepackage{booktabs}
\usepackage{algorithm}
\usepackage{algorithmic}
\usepackage{subcaption}

\def\E{{\rm E}} 
\def\P{{\rm P}}

\def\sp{{\sigma_{\rm p}}}
\def\ia{\Delta I}
\def\ha{\Delta H}
\def\ip{\Delta_{\rm p}I}
\def\hp{\Delta_{\rm p} H}
\def\simp{s_{\rm p}}
\def\qp{q_{\rm p}}

\def\be{\begin{equation}}
\def\ee{\end{equation}}

\newtheorem{theorem}{Theorem}
\newtheorem{cor}{Corollary}
\newtheorem{prop}{Proposition}
\newtheorem{rem}{Remark}

\title{Pairwise Adjusted Mutual Information}

\author{
Denys Lazarenko\\Technische Universität München\\Germany\\{\tt denys.lazarenko@tum.de} \And 
Thomas Bonald\\ Institut Polytechnique de Paris\\France\\ {\tt thomas.bonald@telecom-paris.fr}}

\begin{document}

\maketitle

\begin{abstract}
A well-known metric for quantifying the similarity between two clusterings  is the  adjusted mutual information. Compared to mutual information, a corrective term based on random permutations of the labels is introduced, preventing  two clusterings being similar by chance. Unfortunately, this adjustment makes the metric computationally expensive. 
In this paper, we propose a novel adjustment based on {pairwise} label permutations instead of full label permutations. Specifically, we consider  permutations where only two samples, selected uniformly at random, exchange their labels. We show that the corresponding adjusted metric, which can be expressed explicitly, behaves similarly to the standard adjusted mutual information for assessing the quality of a clustering, while having a much lower  time complexity. Both metrics are compared  in terms of quality and performance on experiments based on synthetic and real data. 
\end{abstract}

\section{Introduction}

A well-known metric for quantifying the similarity between two clusterings  of the same data is the  adjusted mutual information \cite{nguyen2009information,vinh2010information}. 
Compared to mutual information, this metric is  {\it adjusted} against chance, meaning that the similarity   cannot  be due to randomness but only to the structure of the dataset, appearing in both clusterings. This is the reason why this metric is widely used in  unsupervised learning, see \cite{zhang2013general,thirion2014fmri,taha2015metrics,yang2016comparative,wang2017visualization} for various applications.

The standard way of adjusting mutual information against chance  is  through random label permutations of one of the clusterings \cite{vinh2010information}. Unfortunately, this adjustment makes the metric computationally expensive. Specifically, the time complexity of the metric is in $O(\max(k,l)n)$, where $k, l$ are the numbers of clusters in each clustering and $n$ is the number of samples \cite{romano2014standardized}. As a comparison, the time complexity   of mutual information is equal to 
$O(kl)$ given the  {contingency matrix}  of the clusterings, i.e., the matrix counting the number of samples in each cluster pair. 
The additional computational effort required by adjustment is  significant as the number of samples $n$ is typically much larger than the numbers of clusters $k,l$. 

In this paper, we  propose a novel adjustment based on {\it pairwise} permutations. That is, we consider  permutations where only two samples, selected uniformly at random, exchange their labels. We show that the corresponding adjusted metric, we refer to as {\it pairwise adjusted mutual information},  is as efficient as adjusted mutual information for assessing the quality of a clustering,  with a much lower  time complexity. In particular, the time complexity is the  {\it same}  as that of mutual information. The gain in complexity is significant, as the computation time is now independent of the number of samples $n$, given the contingency matrix. 

The rest of the paper is organized as follows. We first provide the  definition and key properties of adjusted mutual information in the general setting of information theory. We then introduce mutual information with pairwise adjustement and explain why the exact same properties are satisfied by this new notion of adjusted mutual information. The application of both notions of adjustment to  clustering, including the explicit expressions of the corresponding metrics, is presented in section \ref{sec:cluster}. Experiments on both synthetic and real data are presented in section \ref{sec:exp}. Section \ref{sec:conc} concludes the paper.

\section{Adjusted mutual information}
\label{sec:ami}

Let $P$ be the uniform probability measure on  $\Omega = \{1,\ldots,n\}$, for some positive integer $n$. Let $X,Y$ be random variables on the probability space $(\Omega, P)$. Without any loss of generality, we assume that  $X$ and $Y$ are mapping from $\Omega$ to sets consisting of  consecutive  integers, starting from 1.
Denoting by $H$ the entropy, the mutual information between $X$ and $Y$ is defined by \cite{cover}:
\be\label{eq:info}
I(X, Y) = H(X) + H(Y) - H(X,Y).
\ee
This is the information shared by $X$ and $Y$, which is equal to 0 if $X$ and $Y$ are independent.
A distance between $X$ and $Y$ can then be defined by:
$$%\be\label{eq:d}
d(X,Y) = H(X,Y) - I(X,Y) = H(X|Y) + H(Y| X).
$$%\ee
This distance, known as the variation of information, is a metric in the quotient space of random variables under the equivalence relation $X\sim Y$ if and only if there is some  bijection $\varphi$ such that $X = \varphi(Y)$ \cite{vi}.

\paragraph{Adjusted mutual information.}
 The adjusted mutual information between $X$ and $Y$, corresponding to  the mutual information between $X$ and $Y$ {\it adjusted} against chance, is defined by:
\be\label{eq:def}
\ia(X,Y) = I(X,Y) - \E(I(X, Y_\sigma)),
 \ee
 where  $Y_\sigma$ is the random variable $Y \circ \sigma$, for any   permutation $\sigma$ of $\{1,\ldots, n\}$, and the expectation is taken over all permutations $\sigma$, chosen uniformly at random. 

 \begin{rem}[Normalization]\label{rem:norm}
 It is frequent to also normalize adjusted mutual information, so as to get a score between 0 and 1 \cite{vinh2010information,romano2014standardized}. In this paper, we only focus on the adjustment step. Note that  normalization can be equally applied to both considered notions of adjustment and thus be studied separately.
 \end{rem}
 
We have the equivalent definition:
\begin{align}
\ia(X,Y) &=\E(H(X, Y_\sigma)) - H(X,Y),\nonumber \\ 
&= \frac 1 2 (\E(d(X, Y_\sigma)) - d(X,Y)).\label{eq:dist}
 \end{align}
 This equivalence follows from
Proposition \ref{prop:equiv} and the fact that  the definition is symmetric in $X$ and $Y$. 
All proofs are deferred to the appendix.

\begin{prop}\label{prop:equiv}
We have for any random variables $X$ and $Y$:
\begin{align*}
H(X)&= \E(H(X_\sigma)),\\
 \E(H(X, Y_\sigma)) &= \E(H(X_\sigma, Y)),\\
 \E(I(X, Y_\sigma)) &= \E(I(X_\sigma, Y)).
 \end{align*}
\end{prop}
 
 In view of \eqref{eq:dist}, we expect $\ia(X,Y)$ to be positive if $X$ and $Y$ share information, as $X$ is expected to be closer to $Y$ (for the distance $d$)  than to $Y_\sigma$, a randomized version of $Y$. 
There are specific cases where $\ia(X,Y)=0$, as stated in 
Proposition  \ref{prop:zero}; these cases will be interpreted  in terms of clustering  in section \ref{sec:cluster}. 

 \begin{prop}\label{prop:zero}
 We have $\ia(X,Y)=0$  whenever $Y$ (or $X$, by symmetry) is constant or equal to some permutation of  $\{1,\ldots,n\}$.
\end{prop}

% Unlike mutual information, the adjusted mutual information can be negative. This is a consequence of Proposition \ref{prop:perm}  below.
% Another key difference is that it depends on both random variables $X, Y$ (as mappings from the probability space) and  not only  on their joint distribution, a property that turns out to be critical in the application to clustering presented in  section \ref{sec:cluster}.
%
% \begin{prop}\label{prop:perm}
%We have for any random variables $X$ and $Y$:
%$$
%\E(\ia(X,Y_\sigma)) = 0.
%$$
%\end{prop}
%\begin{proof}
%This follows from:
%$$
%\E( \ia(X,Y_\sigma)) =\E( I(X,Y_\sigma)) - \E(I(X, Y_{\sigma \circ \sigma'})) = 0.
% $$
%\end{proof}
%
%In words, Proposition \ref{prop:perm} states that the expected adjusted mutual information between $X$ and $Y$  is equal to zero when $Y$ is randomized.  This means in particular that for any random variables 
%$X$ and $Y$ with positive adjusted mutual information, there must exist some permutation $\sigma$ such  that $X$ and $Y_\sigma$ have negative adjusted mutual information.

 \paragraph{Adjusted entropy.} Observing that $H(X) = I(X, X)$, 
we define similarly the adjusted entropy of $X$ by:
$$
 \ha(X) = \ia(X,X) =  H(X) - \E(I(X, X_\sigma)).
 $$
By \eqref{eq:info}, we get:
\be\label{eq:ae2}
  \ha(X) =\E(H(X, X_\sigma)) - H(X)= \frac 12 \E(d(X,X_\sigma)).
 \ee
Since $d$ is a metric, this shows that  the adjusted entropy of $X$ is non-negative. 

\begin{prop}\label{prop:entropy}
We have $\ha(X) = 0$ if and only if  $X$ is constant or equal to some permutation of $\{1,\ldots,n\}$.
\end{prop}

Proposition \ref{prop:entropy} characterizes random variables with zero adjusted entropy. 
Again, this result will be interpreted in terms of clustering in section \ref{sec:cluster}.

\section{Pairwise adjustment}

In this section, we introduce pairwise adjusted mutual information. The definition is the same as adjusted mutual information, except that  the permutation  $\sigma$ is now restricted to the set of pairwise permutations. Specifically, we consider permutations $\sigma$ for which there exists $i,j \in \{1,\ldots,n\}$ such that $\sigma(i) = j$ and $\sigma(j) = i$, whereas $\sigma(t) = t$ for all $t\ne i,j$.
We consider the set of such permutations $\sigma$ where the samples $i,j$ are drawn uniformly at random in the set $\{1,\ldots,n\}$. We denote by $\sp$ such a random permutation. Observe that $\sp$ is the identity with probability $1/n$ (the probability that $i=j$). 

\paragraph{Pairwise adjusted mutual information.}
We  define the {\it pairwise adjusted mutual information} as:
$$
\ip(X,Y) = I(X,Y) - \E(I(X, Y_\sp)).
$$
This is exactly the same definition as the adjusted mutual information, except for the considered permutations $\sp$. It can be readily verified that the same properties  apply, with  the exact same proofs, a key property being that the random permutations $\sp$ and $\sp^{-1}$ have the same distributions.
In particular, we  have the analogue of \eqref{eq:dist}:
 \begin{align}
\ip(X,Y)& =\E(H(X, Y_\sp)) - H(X,Y),\nonumber\\
&= \frac 1 2 (\E(d(X, Y_\sp)) - d(X,Y)). \label{eq:ip2} 
 \end{align}
Moreover, $\ip(X,Y)=0$ whenever $X$ or $Y$ is 
constant or equal to some permutation of $\{1,\ldots,n\}$.

\paragraph{Pairwise adjusted entropy.}
We  also define the {\it pairwise adjusted entropy} as:
$$
 \hp(X) = \ip(X,X) =  H(X) - \E(I(X, X_\sp)).
 $$
 We have $\hp(X) \ge 0$, with equality if and only if  $X$ is constant or equal to some permutation of $\{1,\ldots,n\}$.

\section{Application to clustering}
\label{sec:cluster}

Let $A = \{A_1,\ldots,A_k\}$ and $B= \{B_1,\ldots,B_l\}$  be two partitions of some finite set $\{1,\ldots,n\}$ into $k$ and $l$ clusters, respectively. 
Let  $\Omega = \{1,\ldots,n\}$ and $\P$ be  the uniform probability measure over $\Omega$. 
Consider the  random variables $X$ and $Y$ defined on $(\Omega, \P)$ by $X^{-1}(i) = A_i$ for all $i=1,\ldots,k$ and 
$Y^{-1}(j) = B_j$ for all $j=1,\ldots,l$. Note that $X(\omega)$ and $Y(\omega)$ can be interpreted as the {\it labels} $i$ and $j$ of sample $\omega$ in clusterings $A$ and $B$, for each $\omega\in \{1,\ldots,n\}$.

We denote by $a_i = |A_i|$ the size of cluster $A_i$, by $b_j = |B_j|$ the size of cluster $B_j$, and by $n_{ij} = |A_i \cap B_j|$ the number of samples both in cluster $A_i$ and  cluster $B_j$, for all $i=1,\ldots,k$ and $j=1,\ldots,l$. The matrix $(n_{ij})_{1\le i\le k, 1\le j\le l}$ is known as the {\it contingency matrix}. Note that $a_i$ and $b_j$ are  the respective sums of row $i$  and column $j$ of the contingency matrix.

\paragraph{Adjusted mutual information.} A well-known metric for assessing the  
  similarity $s(A,B)$ between clusterings $A$ and $B$ is  the adjusted mutual information\footnote{Recall that we don't normalize the metric, see Remark \ref{rem:norm}.}  $\ia(X,Y)$ between the corresponding random variables $X$ and $Y$. 
In words, this is the common information shared by  clusterings $A$ and $B$ not due to randomness. 

By Proposition \ref{prop:zero}, we have $s(A,B)=0$ whenever clustering $A$ (or $B$, by symmetry) is trivial, that is, it consists of  a single cluster or of $n$ clusters (one per sample). This is a key property, showing the interest of the adjustment.

It is known that \cite{vinh2010information}:
 \begin{align}\label{eq:ami}
  \begin{split}
&s(A,B) = -\sum_{i=1}^k\sum_{j=1}^l \frac {n_{ij}}n \log \frac {n_{ij}}n + \sum_{i=1}^k\sum_{j=1}^l \sum_{c = (a_i+b_j - n)^+}^{\min(a_i, b_j)} \\
 &\frac{a_i!b_j!(n-a_i)!(n-b_j)!}
{n!c!(a_i-c)!(b_j-c)!(n-a_i-b_j+c)!}\frac {c}n \log \frac {c}n,
\end{split}
\end{align}
with the notation $(\cdot)^+ = \max(\cdot, 0)$.
The time complexity of this formula, which is dominated by the second term, is in $O(\max(k,l)n)$ \cite{romano2014standardized}. In particular, it is linear in the number of samples $n$.

 Interestingly, we can similarly assess the quantity of information $q(A)$ contained  in  clustering $A$ through the adjusted entropy $\ha(X)$ of the corresponding random variable $X$. This is the  information contained in $A$  not due to randomness. 
 We have $q(A) \ge 0$ and, 
 by Proposition \ref{prop:entropy}, $q(A) = 0$  if and only if clustering $A$ is trivial, that is, it consists of  a single cluster or of $n$ clusters (one per sample).

Since $q(A) = s(A, A)$, it follows from  \eqref{eq:ami} that:
 \begin{align*}%\label{eq:q}
 \begin{split}
q(A)&= -\sum_{i=1}^k \frac {a_i}n \log \frac {a_i}n + \sum_{i,j=1}^K \sum_{c = (a_i+a_j-n)^+}^{\min(a_i, a_j)} \\
&\frac{a_i!a_j!(n-a_i)!(n-a_j)!}
{n!c!(a_i-c)!(a_j-c)!(n-a_i-a_j+k)!}\frac {c}n \log \frac {c}n.
\end{split}
\end{align*}
The time complexity of this formula, also dominated by the second term, is in $O(kn)$. Again,  this complexity is linear in    the number of samples $n$. %, even if the cluster sizes $a_1,\ldots,a_k$ are given.

\paragraph{Pairwise adjusted mutual information.} The main contribution of the paper is the following new measure of 
  similarity $\simp(A,B)$ between clusterings $A$ and $B$, based on  the pairwise adjusted mutual information $\ip(X,Y)$ between the corresponding random variables $X$ and $Y$. 
We have an explicit expression for this similarity:

\begin{theorem}\label{theo:pairsim}
We have:
\begin{align}\label{eq:pami}
\begin{split}
\simp(A,B) &=2 \sum_{i=1}^k \sum_{j=1}^l \frac{n_{ij}(n -a_i -b_j + n_{ij})}{n^2}\\
&\times  \left(\frac{n_{ij}} n \log\frac{n_{ij}}n - \frac{n_{ij} - 1} n \log\frac{n_{ij} - 1}n\right)\\
& + 2\sum_{i=1}^k \sum_{j=1}^l \frac{(a_i - n_{ij})(b_j - n_{ij})}{n^2}\\
&\times\left(\frac{n_{ij}} n \log\frac{n_{ij}}n - \frac{n_{ij} + 1} n \log\frac{n_{ij} + 1}n \right).
\end{split}
\end{align}
\end{theorem}
The time complexity of this formula is  in $O(kl)$, like mutual information. It is independent of the number of samples $n$, given the contingency matrix. Corollary \ref{cor:pairsim-sparse} shows that 
the time complexity reduces to $O(m)$ the number of non-zero entries of the contingency matrix, provided the latter is stored in sparse format.

\begin{cor}\label{cor:pairsim-sparse}
We have:
\begin{align*}%\label{eq:pami-sparse}
\begin{split}
\simp(A,B) &= 2\sum_{i, j: n_{ij} > 0} \frac{n_{ij}(n -a_i -b_j + n_{ij})}{n^2}\\
&\times  \left(\frac{n_{ij}} n \log\frac{n_{ij}}n - \frac{n_{ij} - 1} n \log\frac{n_{ij} - 1}n\right)\\
& + 2\sum_{i, j: n_{ij} > 0} \frac{(a_i - n_{ij})(b_j - n_{ij})}{n^2}\\
&\times\left(\frac{n_{ij}} n \log\frac{n_{ij}}n - \frac{n_{ij} + 1} n \log\frac{n_{ij} + 1}n  +\frac 1 n \log \frac 1 n \right)\\
& -  2 \left(n^2 - \sum_{i=1}^k  a_i^2 -  \sum_{j=1}^l  b_i^2 + \sum_{i, j: n_{ij} > 0} n_{ij}^2\right)\\
&\times \frac 1 n \log \frac 1 n .
\end{split}
\end{align*}
\end{cor}

Similarly, we can define  the quantity of  information $\qp(A)$ in clustering $A$ through the  pairwise adjusted entropy $\hp(X)$ of the corresponding random variable $X$. Again, $\qp(A) \ge 0$, with  $\qp(A) = 0$  if and only if clustering $A$ is trivial.

\begin{cor}\label{cor:pairq}
We have:
\begin{align*}%\label{eq:qp}
\begin{split}
\qp(A) &= 2\sum_{i=1}^k \frac{a_i(n -a_i)}{n^2}\\
&\times \left(\frac{a_{i}} n \log\frac{a_{i}}n - \frac{a_{i} - 1} n \log\frac{a_{i} - 1}n - \frac 1 n \log \frac 1 n \right).
\end{split}
\end{align*}
\end{cor}

Note that the time complexity of this formula in $O(k)$. It only depends on the number of clusters $k$, and not on the number of samples $n$.

\section{Experiments}
\label{sec:exp}

In this section, we compare both notions of adjusted mutual information through experiments involving synthetic and real data.
The experiments are run on a computer equipped with an AMD Ryzen Threadripper 1950X 16-Core Processor and 32 GB of RAM, with a a Debian 10 OS.
All codes and datasets used in the experiments are available online\footnote{See \url{https://github.com/denyslazarenko/Pairwise-Adjusted-Mutual-Information}}. 

\paragraph{Synthetic data.} We start with the simple case of 
 $n=100$ samples with clusters  of even sizes, consisting of consecutive samples. Specifically, we consider the set of clusterings $A^{(s)}$, consisting of clusters of size $s$ (except possibly the last one), for $s=1, 2, \ldots, 100$. In particular, both $A^{(1)}$ and $A^{(100)}$ are trivial clusterings while $A^{(5)}$ consists of 20 clusters of size 5. 
  
\begin{figure}[h]
     \centering
     \begin{subfigure}{0.45\textwidth}
         \centering
         \includegraphics[width=7cm]{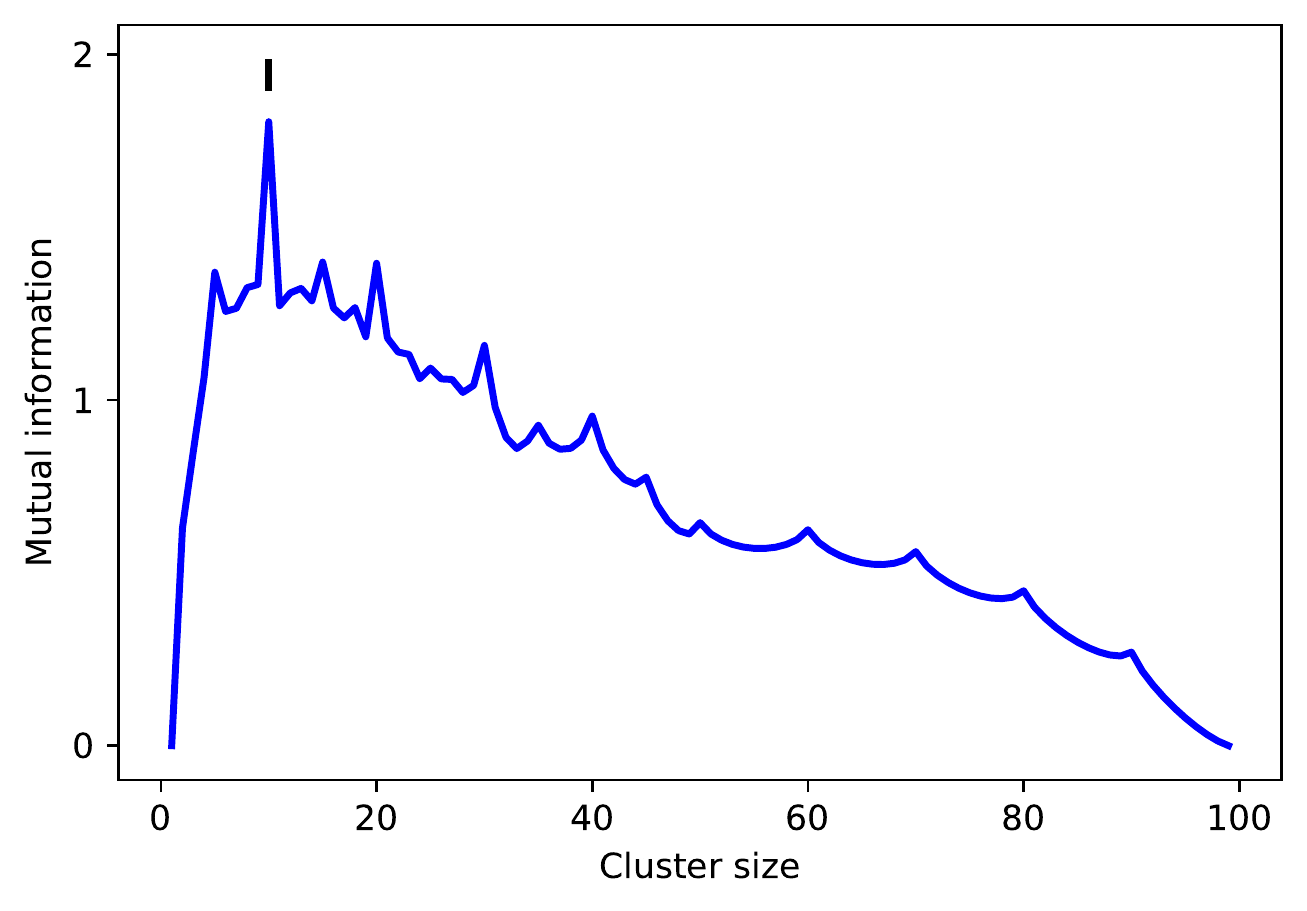}
         \caption{Full adjustment}
     \end{subfigure}
     \begin{subfigure}{0.45\textwidth}
         \centering
         \includegraphics[width=7cm]{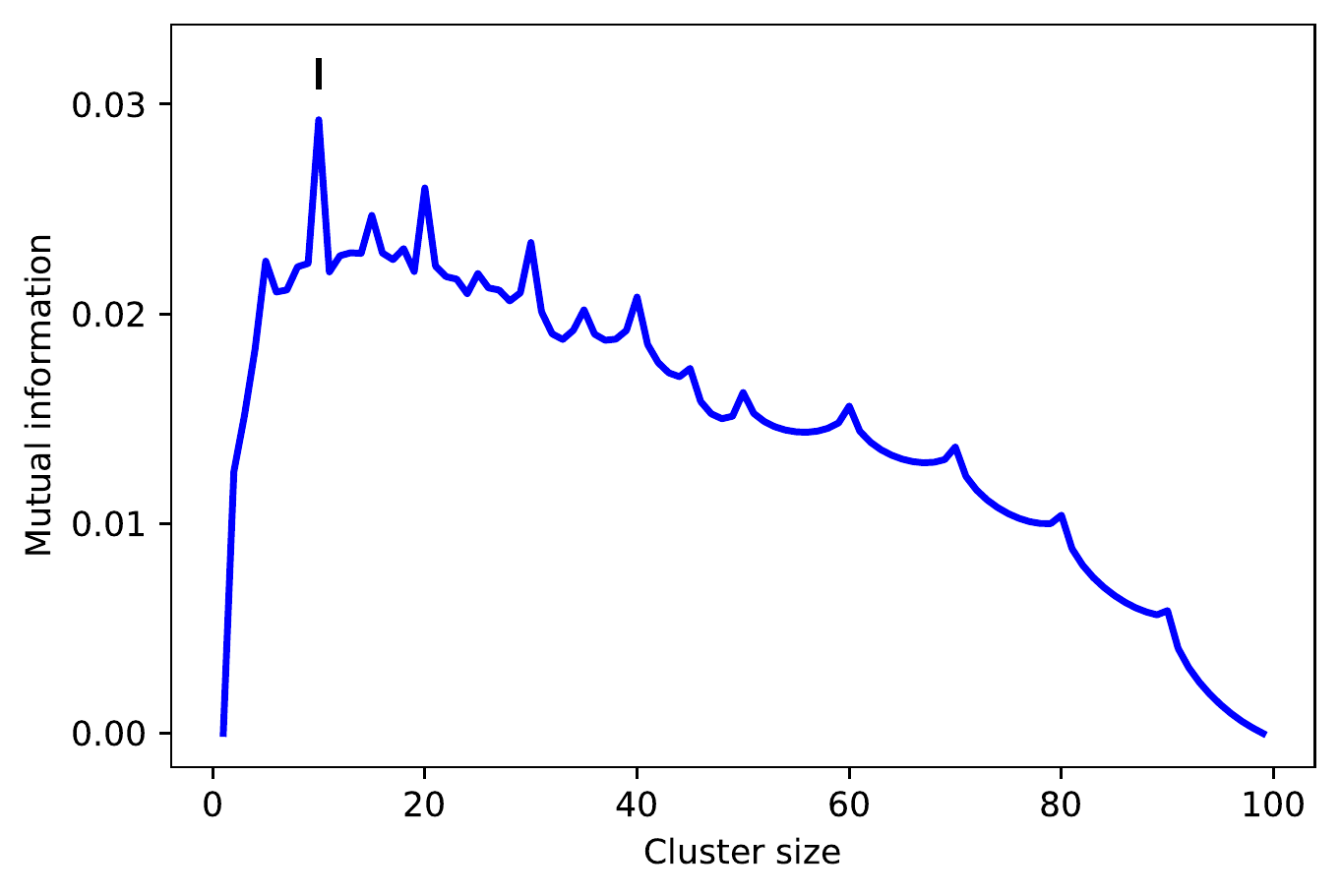}
         \caption{Pairwise adjustment}
     \end{subfigure}
     \caption{Comparison of metrics on synthetic data ($n=100$).\label{fig:res}}
\end{figure}

 Figure \ref{fig:res} gives the   similarity  between clusterings  $A^{(10)}$ and $A^{(s)}$  with respect to $s$ in terms of adjusted mutual information, for both notions of adjustment. We observe very close behaviors, suggesting that both notions of adjustment tend to capture the same patterns in the clusterings.  Note that the maximum similarity is attained for $s=10$ in both cases, as expected. The similarity is equal to 0 for $s\in \{1,100\}$ for both cases, in agreement with 
    Proposition \ref{prop:zero}. 
  We  also observe local peaks at $s = 20, 30,\ldots, 90$, which can be interpreted by the fact that clustering  $A^{(10)}$ is  a refinement of clustering $A^{(s)}$ for these values of $s$; similarly, the local peak  at $s=5$ may be interpreted by the fact that clustering $A^{(5)}$ is a refinement of clustering $A^{(10)}$.

 We now consider random clusterings. Specifically, we assign  $n$ samples to $k$ clusters independently at random, according to some probability distribution $p = (p_1,\ldots,p_k)$, which is itself drawn at random\footnote{Namely, $p \propto U$ where $U = (U_1,\ldots,U_k)$ is a vector of $k$ i.i.d.~random variables uniformly distributed over $[0,1]$.}. Consider three such random clusterings 
 $A$, $B$, $C$ (with the same parameters $n$ and $k$, but different probability distributions $p$). We would like to know whether $A$ is ``closer" to  $B$ or to $C$. In particular, we are interested in   testing whether both notions of adjusted mutual information give the same ordering in the sense that:
 \be\label{eq:order}
(s(A, B) - s(A, C)) (\simp(A, B) - \simp(A, C)) \ge 0.
\ee
We compute the average precision score (fraction of triplets $A, B, C$ for which \eqref{eq:order} is true) over 1\,000 independent samples of $A,B,C$, for different values of $n$ and $k$. We repeat the experiment 100 times to get the mean and standard deviation. The results are given in Table \ref{tab:prec}. 
We observe a very high precision score, always higher than $93\%$, showing that both notions of adjusted mutual information tend to give the same ordering of these random clusterings.

\begin{table}[h]
\centering
\begin{tabular}{ccc}
\toprule
$n$ & $k$  &Precision score \\
\midrule
100 & 2 & $0.972\pm 0.004$     \\
 100 &5     & $0.952\pm 0.007$     \\
 100 & 10 & $0.943 \pm 0.006$     \\
 100 & 20 & $0.955\pm 0.008$\\
500 & 20 & $0.936\pm 0.007$     \\
 1000 &20     & $0.933\pm 0.006$     \\
 1000 & 50 & $0.949 \pm 0.008$     \\
\bottomrule
\end{tabular}
\caption{Precision score (mean $\pm$ standard deviation)}
\label{tab:prec}
\end{table}

\begin{figure}[h]
     \centering
           \includegraphics[width=7cm]{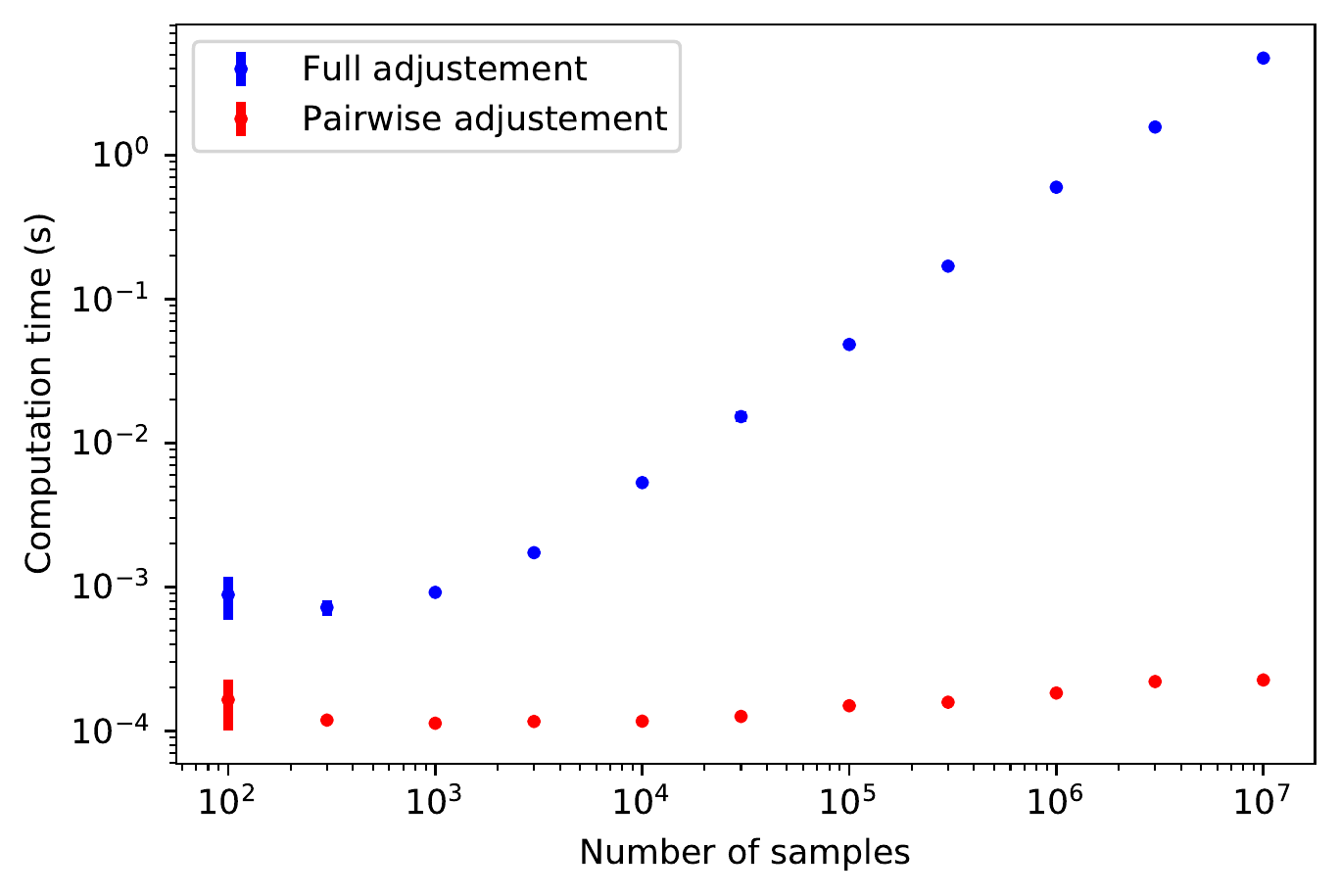}
         \caption{Computation time with respect to    $n$ (mean $\pm$ standard deviation).\label{fig:time}}
\end{figure}

For the performance gain, we compare the computation times of both versions of adjusted mutual information for the similarity between clusterings $A$ and $B$, where $A$ consists of   $k=10$ clusters of  same size and $B$ is a random clustering, drawn as in the previous experiment. 
Both versions of adjusted mutual information are coded in Python, with the standard version imported from \href{http://scikit-learn}{scikit-learn}. 
 Figure \ref{fig:time} shows the computation time when the number of samples $n$ grows from $10^2$  to $10^7$. The  performance gain brought by pairwise adjustement is significant. In particular, the computation time becomes independent of the number of samples.

\begin{figure}[h]
     \centering
     \begin{subfigure}{0.45\textwidth}
         \centering
         \includegraphics[width=7cm]{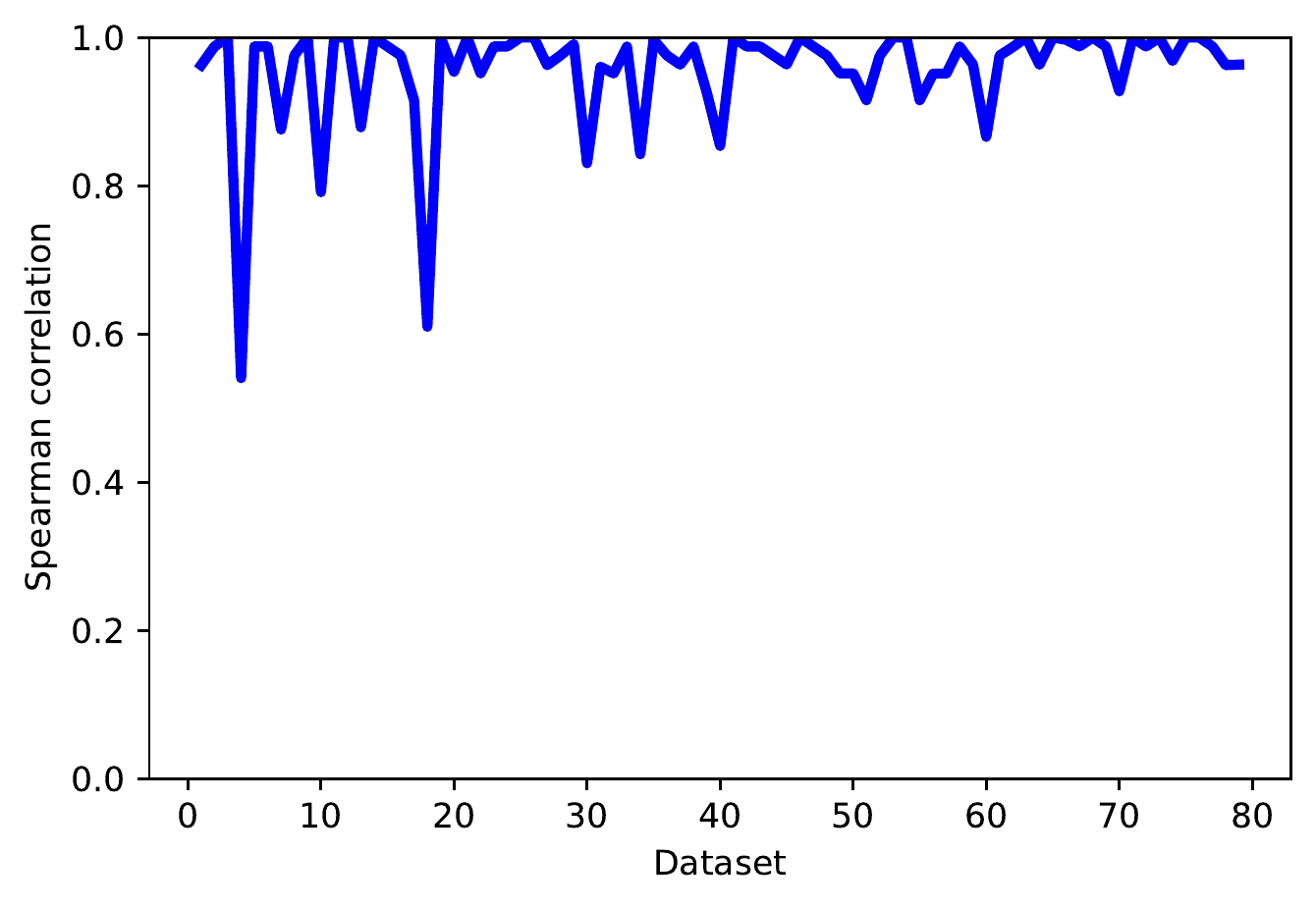}
         \caption{Spearman correlation.}
     \end{subfigure}
     \begin{subfigure}{0.45\textwidth}
         \centering
         \includegraphics[width=7cm]{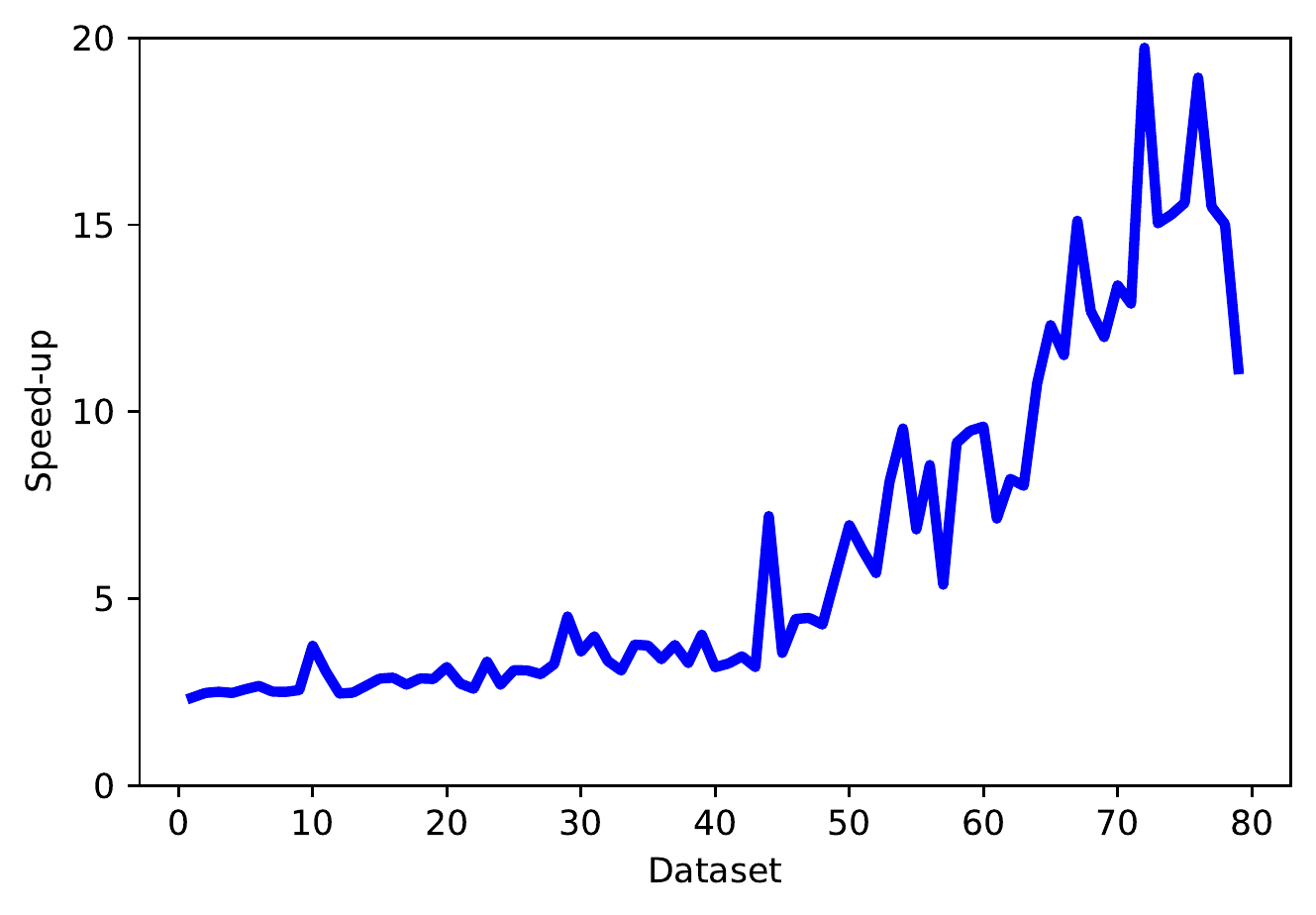}
         \caption{Speed-up of pairwise adjustment.}
     \end{subfigure}
     \caption{Comparison of metrics on real data.\label{fig:real_data}}
\end{figure}

\paragraph{Real data.} 
For real data, we consider the 79 datasets of the benchmark suite \cite{clustering_benchmarks}\footnote{See \url{https://github.com/gagolews/clustering_benchmarks_v1}.}. We apply to each dataset each of the 
following clustering algorithms:
\begin{enumerate}
\item $k$-means
\item Affinity propagation
\item Mean shift
\item Spectral clustering
\item Ward
\item Agglomerative clustering
\item DBSCAN
\item OPTICS
\item Birch
\item Gaussian Mixture
\end{enumerate}
We use  the scikit-learn implementation of these algorithms, with the corresponding default parameters\footnote{See \url{https://scikit-learn.org/stable/auto_examples/cluster/plot_cluster_comparison.html}.}\footnote{Dimension reduction is applied to the MNIST datasets, consisting of 70\,000 images of size  $28\times 28$ each, see the supplementary material for details.}. We get 10 clusterings per dataset. The quality of each clustering is assessed through the similarity with the available ground-truth labels, using adjusted mutual information with either full adjustment or pairwise adjustment. We then compute  the Spearman correlation of the corresponding similarities, a value of 1 meaning the exact same ordering of the 10 clusterings with full adjustment and pairwise adjustment.
The results are shown in Figure \ref{fig:real_data}, together with the speed-up in computation time due to pairwise adjustment.  In both cases, the 79 datasets are ordered by the number of samples, ranging from 105 to 105\,600 \cite{clustering_benchmarks}.

We first observe that the correlation is very high, suggesting again that both notions of adjusted mutual  information tend to provide the same results. For 65 datasets among 79, the Spearman correlation is higher than 95\%.  %On average, the Spearman correlation  is equal to 0.96, with a standard deviation of 0.07. 
As for the computation time, we observe a significant performance gain, by one order of magnitude  for the largest datasets.

\section{Conclusion}
\label{sec:conc}

We have proposed another way of adjusting mutual information against chance, through pairwise label permutations. The novel metric, whose explicit expression is given in Theorem \ref{theo:pairsim}, has a much lower complexity than the usual adjusted mutual information. 
Interestingly, both metrics can also  be used to assess the quantity of information contained in a clustering, which the common property of being    equal to 0 if and only if the clustering is trivial, as stated in Proposition \ref{prop:entropy}; again, the pairwise adjusted entropy, given in Corollary \ref{cor:pairq}, has a much lower complexity.
Experiments on  synthetic and real data show that pairwise adjusted mutual information tends to provide the same results as the usual adjusted mutual information for comparing clusterings, while involving much less computations. 

For future work, we plan to extend this idea to other similarity metrics. While the practical interest is less obvious for 
 the Adjusted Rand Index \cite{hubert1985comparing}, due to the fact that the time complexity of this metric is already  independent of the number of samples, it would be worth considering other versions of information theoretic measures, as those studied in  \cite{romano2016adjusting}.

\section*{Appendix}

\section*{Proof of Proposition \ref{prop:equiv}}

The first equality follows from the fact that $X$ and $X_\sigma$ have the same distribution. 
Specifically, we have for any positive integer $k$,
\begin{align*}
 \P(X_\sigma=k) &= \P((X\circ \sigma)^{-1}(k)),\\
& =  \P( \sigma^{-1}(X^{-1}(k))),\\
&=  \P( X^{-1}(k)),\\
& = \P(X = k).
\end{align*}

For the second, we observe that if $\sigma$ is a random permutation of $\{1,\ldots, n\}$, chosen uniformly at random, so is $\sigma^{-1}$ which implies:
$$
\E(H(X, Y_\sigma))  = \E(H(X, Y_{\sigma^{-1}})) =  \E(H(X_\sigma, Y)),
$$
where we have used the first equality and  the fact that:
$$
 (X, Y_{\sigma^{-1}})\circ \sigma = (X_\sigma, Y).
$$
The third equality is a direct consequence of the two first.

\section*{Proof of Proposition \ref{prop:zero}}

If $Y$ is constant, then $Y=Y_\sigma$ for all permutations $\sigma$ and the result follows from \eqref{eq:dist}.
Now assume that $Y$ is a permutation of $\{1,\ldots,n\}$. Then $H(X,Y) = H(Y) = \log(n)$ and $I(X,Y) = H(X)$, for any random variable $X$. It then follows from \eqref{eq:def} (and the symmetry  in $X$ and $Y$) that
$$
\Delta I(X,Y) = I(X, Y) - \E(I(X_\sigma, Y)) = 0.
$$

\section*{Proof of Proposition \ref{prop:entropy}}

 If $\ha(X) = 0$, then $d((X, X_\sigma)) = 0$ for all permutation $\sigma$. In particular, there exists some bijection $f$ such that  $X_\sigma = f(X)$. 
Now assume that for some integer $i$, the event: $$A = \{\omega: X(\omega) = i\}$$ is such that $1 < |A| < n$.  Then there exists some $j\ne i$ such that the event $B =\{\omega: X(\omega) = j\}$ is not empty. Choose $a\in A$ and $b\in B$ and define $\sigma$ as the permutation of $a$ and $b$. Then $X_\sigma(a) = X(b) = j$ while $X_\sigma(a') = X(a') = i$ for all $a'\in A\setminus\{a\}$. So  $X_\sigma(a)  \ne X_\sigma(a')$ while $X(a) = X(a')$ for all $a'\in A\setminus\{a\}$, which contradicts the existence of some mapping $f$  that $X_\sigma = f(X)$. Thus for each integer $i$, the cardinal of the event $A = \{\omega: X(\omega) = i\}$ is $0, 1$ or $n$. This implies that $X$ is constant or equal to some permutation of $\{1,\ldots,n\}$.

\section*{Proof of Theorem \ref{theo:pairsim}}

Consider two  items selected uniformly at random in $\{1,\ldots,n\}$. Let $A_{i_1}, B_{j_1}$ be the clusters of the first item, $A_{i_2}, B_{j_2}$ be the clusters of the second item. In particular, these items belong respectively to the sets $A_{i_1} \cap B_{j_1}$ and $A_{i_2} \cap B_{j_2}$. The probability of this event is:
$$
\frac{n_{i_1j_1}n_{i_2j_2}}{n^2}.
$$
Now assume that these items exchange their labels for the first clustering, so that the first item move to set $A_{i_2}$ while the second item move to the set $A_{i_1}$.  If $i_1=i_2$ or $j_1= j_2$, the new contingency matrix remains unchanged; now if $i_1\ne i_2$ and $j_1\ne j_2$, the new contingency matrix $n'_{ij}$ remains unchanged except for the following entries:
$$
n'_{ij} = \left\{\begin{array}{ll}
n_{ij} - 1 & \text{for }i,j = i_1, j_1 \text{ and } i_2, j_2,\\
n_{ij} + 1 & \text{for }i,j = i_1, j_2 \text{ and } i_2, j_1.
\end{array}
\right.
$$
Using \eqref{eq:ip2}, we obtain the  similarity between clusterings $A$ and $B$:
\begin{align*}
\simp(A,B) &=  \sum_{i_1\ne i_2, j_1\ne  j_2} \frac{n_{i_1j_1}n_{i_2j_2}}{n^2} \\
&\times \left(\frac {n_{i_1j_1}}n \log\frac {n_{i_1j_1}}n - \frac {n_{i_1j_1} - 1}n \log\frac {n_{i_1j_1} - 1}n\right.\\
&+ \frac {n_{i_2j_2}}n \log\frac {n_{i_2j_2}}n - \frac {n_{i_2j_2} - 1}n \log\frac {n_{i_2j_2} - 1}n\\
&+ \frac {n_{i_1j_2}}n \log\frac {n_{i_1j_2}}n - \frac {n_{i_1j_2} + 1}n \log\frac {n_{i_1j_2} + 1}n\\
&+ \left.\frac {n_{i_2j_1}}n \log\frac {n_{i_2j_1}}n - \frac {n_{i_2j_1} + 1}n \log\frac {n_{i_2j_1} + 1}n\right),
\end{align*}
where by convention, $x\log x=0$ for any $x\le 0$. Observing that for any given $i_1, j_1$,
$$
 \sum_{i_2\ne i_1, j_2\ne j_1} n_{i_1j_1}n_{i_2j_2} = n_{i_1j_1}(n - a_{i_1} - b_{j_1} + n_{i_1j_1}),
$$
while for any given $i_1, j_2$,
$$
 \sum_{i_2\ne i_1, j_1\ne j_2} n_{i_1j_1}n_{i_2j_2} = (a_{i_1} - n_{i_1j_2})(b_{j_2} - n_{i_1j_2}),
$$
we get by symmetry:
\begin{align*}
\simp(A,B) &=  2 \sum_{i,j}  \frac{n_{ij}(n -a_i -b_j + n_{ij})}{n^2}\\
&\times \left(\frac{n_{ij}} n \log\frac{n_{ij}}n - \frac{n_{ij} - 1} n \log\frac{n_{ij} - 1}n\right)\\
& + 2 \sum_{i,j}  \frac{(a_i - n_{ij})(b_j - n_{ij})}{n^2}\\
&\times  \left(\frac{n_{ij}} n \log\frac{n_{ij}}n - \frac{n_{ij} +1} n \log\frac{n_{ij} + 1}n\right).
\end{align*}

\section*{Proof of Corollary \ref{cor:pairsim-sparse}}

The proof follows on observing that the second sum in \eqref{eq:pami}  can be written:
\begin{align*}
S & \stackrel{d}{=} \sum_{i,j}  \frac{(a_i - n_{ij})(b_j - n_{ij})}{n^2}\\
&\times  \left(\frac{n_{ij}} n \log\frac{n_{ij}}n - \frac{n_{ij} +1} n \log\frac{n_{ij} + 1}n\right)\\
&= \sum_{i,j: n_{ij}> 0} \frac{(a_i - n_{ij})(b_j - n_{ij})}{n^2}\\
&\times \left(\frac{n_{ij}} n \log\frac{n_{ij}}n - \frac{n_{ij} + 1} n \log\frac{n_{ij} + 1}n \right)\\
&-  \sum_{i,j: n_{ij} = 0} \frac{(a_i - n_{ij})(b_j - n_{ij})}{n^2}\frac{1} n \log\frac{1}n.
\end{align*}
Since 
\begin{align*}
\sum_{i,j: n_{ij} = 0}& {(a_i - n_{ij})(b_j - n_{ij})} =\\
& \sum_{i,j} (a_i - n_{ij})(b_j - n_{ij}) \\
&- \sum_{i,j: n_{ij} > 0} {(a_i-n_{ij})(b_j - n_{ij})}\\
& = n^2 - \sum_{i} a_i^2 -  \sum_{j} b_j^2 \\
&+ \sum_{i,j}n_{ij}^2  -  \sum_{i,j: n_{ij} > 0} {(a_i-n_{ij})(b_j - n_{ij})},
\end{align*}
we get:
\begin{align*}
S &= \sum_{i,j: n_{ij}> 0} \frac{(a_i - n_{ij})(b_j - n_{ij})}{n^2}\\
&\times \left(\frac{n_{ij}} n \log\frac{n_{ij}}n - \frac{n_{ij} + 1} n \log\frac{n_{ij} + 1}n  + \frac{1} n \log\frac{1}n\right)\\
& - \left(n^2 - \sum_{i} a_i^2 -  \sum_{j} b_j^2 + \sum_{i,j}n_{ij}^2\right) \frac{1} n \log\frac{1}n.
\end{align*}

\section*{Proof of Corollary \ref{cor:pairq}}

The proof follows from \eqref{eq:pami} applied to the diagonal contingency matrix
$
n_{ij} = a_i \delta_{ij} = b_j \delta_{ij},
$
where $\delta_{ij}$ denotes the Kronecker symbol.

\bibliographystyle{abbrv}
\bibliography{information}

\end{document}